\documentclass{article}

\usepackage[final]{neurips_2025}

\usepackage[utf8]{inputenc} 
\usepackage[T1]{fontenc}    
\usepackage{hyperref}       
\usepackage{url}            
\usepackage{booktabs}       
\usepackage{amsfonts}       
\usepackage{nicefrac}       
\usepackage{microtype}      
\usepackage{xcolor}         

\usepackage{times}
\usepackage{epsfig}
\usepackage{graphicx}
\usepackage{amsmath}
\usepackage{amssymb}
\usepackage{xcolor}         
\usepackage{float}
\usepackage{caption}
\usepackage{amsfonts,amssymb}
\usepackage{mathrsfs}
\usepackage{tabularx}
\usepackage{wrapfig}
\usepackage{verbatim}
\usepackage{dsfont}
\usepackage{tablefootnote}
\usepackage[stable]{footmisc}
\usepackage{booktabs}
\usepackage{multicol}
\usepackage{multirow}
\usepackage{color}
\usepackage{subcaption}
\usepackage{pythonhighlight}
\usepackage{algpseudocode}
\usepackage{algorithm}
\usepackage{physics}

\renewcommand{\vec}[1]{\mathbf{#1}}

\renewcommand{\paragraph}[1]{\smallskip\noindent\textbf{#1}}

\definecolor{sim}{RGB}{128,0,128}
\definecolor{interp}{RGB}{255, 169, 0}
\definecolor{mytbcol}{RGB}{175,227,246}

\newcommand{\refSec}[1]{Sec.~\ref{sec:#1}}
\newcommand{\refFig}[1]{Fig.~\ref{fig:#1}}

\usepackage{xcolor}
\usepackage{color, colortbl}
\usepackage{xcolor}
\usepackage{adjustbox}
\newcommand{\jiapeng}[1]{#1}

\renewcommand{\vec}[1]{\mathbf{#1}}
\renewcommand{\paragraph}[1]{\smallskip\noindent\textbf{#1}}

\definecolor{tabfirst}{rgb}{1, 0.7, 0.7} 
\definecolor{tabsecond}{rgb}{1, 0.85, 0.7} 
\definecolor{tabthird}{rgb}{1, 1, 0.7} 

\newcommand{\tabfirst}[1]{\colorbox{tabfirst}  {#1}}
\newcommand{\tabsecond}[1]{\colorbox{tabsecond} {#1}}

\usepackage{hyperref}
\hypersetup{
    colorlinks=true,
    linkcolor=black,
    citecolor=green,
    urlcolor=blue
}

\title{ROGR: Relightable 3D Objects 
    using \\ Generative Relighting}
\author{
  Jiapeng Tang$^{1,3}$\thanks{Equal contribution.} \quad 
  Matthew Levine$^1$\footnotemark[1] \quad 
  Dor Verbin$^2$ \quad 
  Stephan J. Garbin$^1$ \\
  \textbf{Matthias Nie{\ss}ner$^3$ \quad 
  Ricardo Martin-Brualla$^1$ \quad 
  Pratul P. Srinivasan$^2$ \quad 
  Philipp Henzler$^1$} \\
  $^{1}$ Google Research \quad
  $^{2}$ Google Deepmind \quad 
  $^{3}$ Technical University of Munich
}

\begin{document}

\maketitle
\begin{figure}[h]
    \vspace{-6mm}
    \centering
    \includegraphics[width=\linewidth]{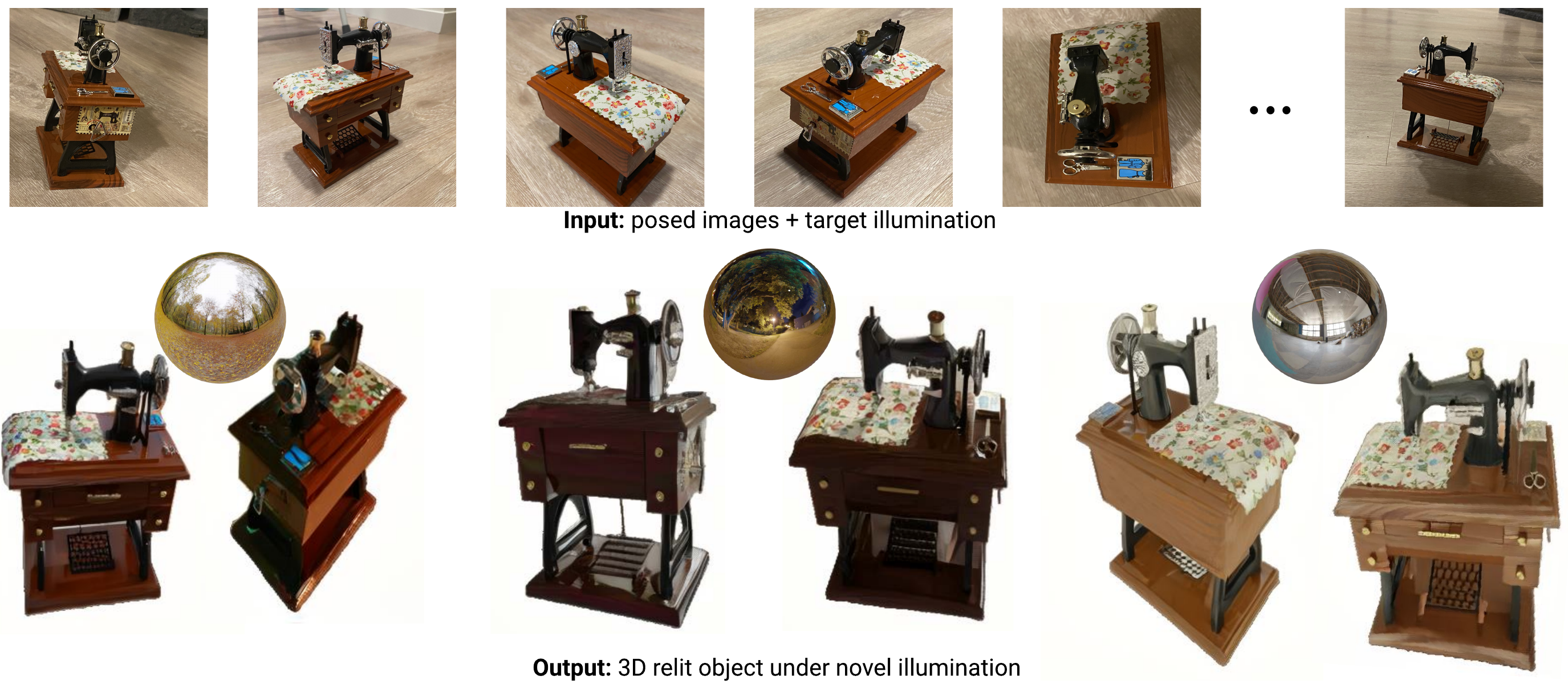}
    \caption{
      Given a set of posed images under unknown illumination (top), our method reconstructs a relightable neural radiance field (bottom), that can be rendered under any novel environment map without further optimization, on-the-fly relighting and novel view synthesis. 
    }
    \label{fig:Teaser}
    \vspace{-2mm}
\end{figure}

\begin{abstract}
We introduce ROGR, a novel approach that reconstructs a relightable 3D model of an object captured from multiple views, driven by a generative relighting model that simulates the effects of placing the object under novel environment illuminations. Our method samples the appearance of the object under multiple lighting environments, creating a dataset that is used to train a lighting-conditioned Neural Radiance Field (NeRF) that outputs the object's appearance under any input environmental lighting. 
The lighting-conditioned NeRF uses a novel dual-branch architecture to encode the general lighting effects and specularities separately.
The optimized lighting-conditioned NeRF enables efficient feed-forward relighting under arbitrary environment maps without requiring per-illumination optimization or light transport simulation. We evaluate our approach on the established TensoIR and Stanford-ORB datasets, where it improves upon the state-of-the-art on most metrics,  and showcase our approach on real-world object captures.
\href{https://tangjiapeng.github.io/ROGR}{Project Page}

\end{abstract}    

\section{Introduction}
\label{SecIntro}

Inserting real-world objects into new environments is a long-standing problem in computer graphics~\cite{debevec2008rendering, debevec2000acquiring}, with numerous applications in the movie and gaming industries. 
While recent years have seen significant progress in 3D object reconstruction for view synthesis using radiance fields~\cite{Nerf,GaussianSplatting}, these techniques represent an object illuminated by a single fixed environment and they do not enable changing the appearance of the object due to changes in the lighting. This work extends 3D object reconstruction to enable rendering the object under arbitrary illumination.

A typical approach for reconstructing relightable 3D representations from images is inverse rendering: optimizing the material and lighting parameters that together explain the captured images. This is particularly challenging for real-world captures as it is brittle and sensitive to mismatches between the real world's physical light transport and the simplified lighting and material models used during optimization. Furthermore, due to the problem's inherent ambiguities, object properties recovered by inverse rendering often appear implausible and unrealistic when viewed under novel lighting. 



At the same time, recent relighting diffusion models~\cite{zeng2024dilightnet,jin2024neural_gaffer,zhao2024illuminerf} have demonstrated impressive capabilities for generating realistic images of objects under arbitrary illumination. However, they only generate a single relit image at a time, which results in inconsistent relighting results when applied to a sequence of viewpoints. While these inconsistently-relit samples can be reconstructed into a single 3D representation~\cite{zhao2024illuminerf}, optimizing a new 3D representation for each new target lighting is tedious and precludes interactive use cases.



We propose a strategy for distilling samples from a relighting diffusion model into a relightable 3D Neural Radiance Field (NeRF) that can be rendered from arbitrary viewpoints under arbitrary novel environment illumination. Given images of an object, we first use a multi-view diffusion network to generate view-consistent relit images under a wide array of environment illuminations. We then use these multi-view multi-illumination samples to train a novel NeRF architecture that predicts outgoing view-dependent color conditioned on a target illumination. 


We evaluate the efficiacy of our method on the task of 3D relighting using both synthetic and real-world benchmarks. Our approach achieves state-of-the-art results but also demonstrates significantly improved test-time performance compared to prior work. These performance gains are due to our generalizable feed-forward relightable NeRF.
\section{Related Work}
\label{SecRelated}
\paragraph{Inverse Rendering for Relighting.}
Recovering relightable 3D representations of objects from images is a longstanding goal in computer vision and graphics. The prevalent approach is inverse rendering: decomposing the object's appearance into its underlying geometry, illumination, and material parameters, and relighting the object by simulating the physical interaction of a new target illumination with the recovered geometry and materials~\cite{ramamoorthi01,yu99, sato1997object, lensch2003image}. 

Modern methods for reconstructing relightable 3D representations with inverse rendering use differentiable rendering techniques~\cite{Li:2018:DMC,NimierDavidVicini2019Mitsuba2} to optimize mesh~\cite{hasselgren2022nvdiffrecmc,NVDiffRec}, distance field~\cite{iron-2022}, or volumetric~\cite{nrf,nerv2021,boss2021nerd,nerfactor,mai2023neural,tensoir,attal2024flash,feng2024arm} representations of object geometry and material parameters. 
While these inverse rendering approaches can be effective, they come with significant limitations: errors in estimating an object's geometry and materials can produce unrealistic appearance when the object is relit under a new illumination and physically-accurate relighting involves computationally-expensive Monte Carlo simulation of global illumination, which can be too slow for interactive use cases.

\paragraph{Precomputed Radiance Transfer and Light Stages.}
Early approaches for interactive relighting in the field of computer graphics proposed the idea of Precomputed Radiance Transfer (PRT)~\cite{sloan02}. As appearance behaves linearly with respect to lighting, an object or scene's appearance can be precomputed under a set of basis lightings and these can be linearly combined to produce appearance under any desired target lighting condition. While PRT-based techniques enable interactive relighting, they struggle with the memory demands of storing the precomputed transfer matrices. To enable rendering under novel illuminations from arbitrary viewpoints, PRT-based methods need to store a full transfer matrix for each 3D point in the scene. These memory requirements are prohibitive even for modestly sized scenes and environment maps, so a large body of work focuses on compressing these PRT matrices~\cite{ng03,sloan03}.

One-light-at-a-time (OLAT) captures in light stages~\cite{debevec00} can be thought of as directly capturing basis vectors of a real object's radiance transfer matrices. Since each image captured by the light stage is of the object illuminated by a single element of a standard lighting basis, they can be linearly combined to reproduce the object's appearance under any desired environment illumination. 

\paragraph{Direct Relighting without Inverse Rendering.}
In the deep learning era, many methods have trained networks to directly output relit images. Early techniques utilized standard convolutional neural networks~\cite{xu2018deep,sun2019single} and more recent approaches are based on powerful diffusion models~\cite{zeng2024dilightnet,zhao2024illuminerf,jin2024neural_gaffer,poirier2024diffusion}. 
These generative relighting models have produced impressive results for image relighting, but they cannot be easily used for 3D relighting as the relit appearance is often not consistent across views. 
IllumiNeRF~\cite{zhao2024illuminerf}, MIS~\cite{poirier2024diffusion}, and Neural Gaffer~\cite{jin2024neural_gaffer} perform 3D relighting by reconciling samples from a 2D image relighting diffusion model into a single NeRF. However, the relit 3D representation needs to be optimized separately for each novel target illumination. Recent work on 3D reconstruction from images of an object with varying illumination~\cite{alzayer2024generative} uses a multi-view diffusion model to relight them to have consistent illumination. However, this approach does not allow relighting using a new unseen illumination condition, so it does not allow for generalization. And it does not utilize the high-frequency details in environment maps.  Zeng et al.~\cite{zeng2023relighting} and RNG~\cite{fan2024rng} utilize shadow and highlight cues as conditioning signals for NeRF. In contrast, we explicitly
use the encoded features of incident light from the specular reflection direction. Additionally, in contrast to all of these approaches, our proposed method trains a generalizable relightable NeRF that can be rendered under any target illumination without additional optimization. 
A concurrent work, RelitLRM~\cite{zhang2024relitlrm}, directly recovers 3D geometry and appearance but does not guarantee consistent geometry across environment maps. In contrast, our method reconstructs constant geometry under varying illumination. DiffusionRenderer~\cite{liang2025diffusion} employs a video diffusion model for relighting but lacks 3D consistency, while our approach enforces it by explicitly modeling geometry and rendering. Moreover, our model enables fast inference, avoiding the long sampling times typical of diffusion-based generation.

\begin{figure*}[t]
\centering
\includegraphics[width=\linewidth]{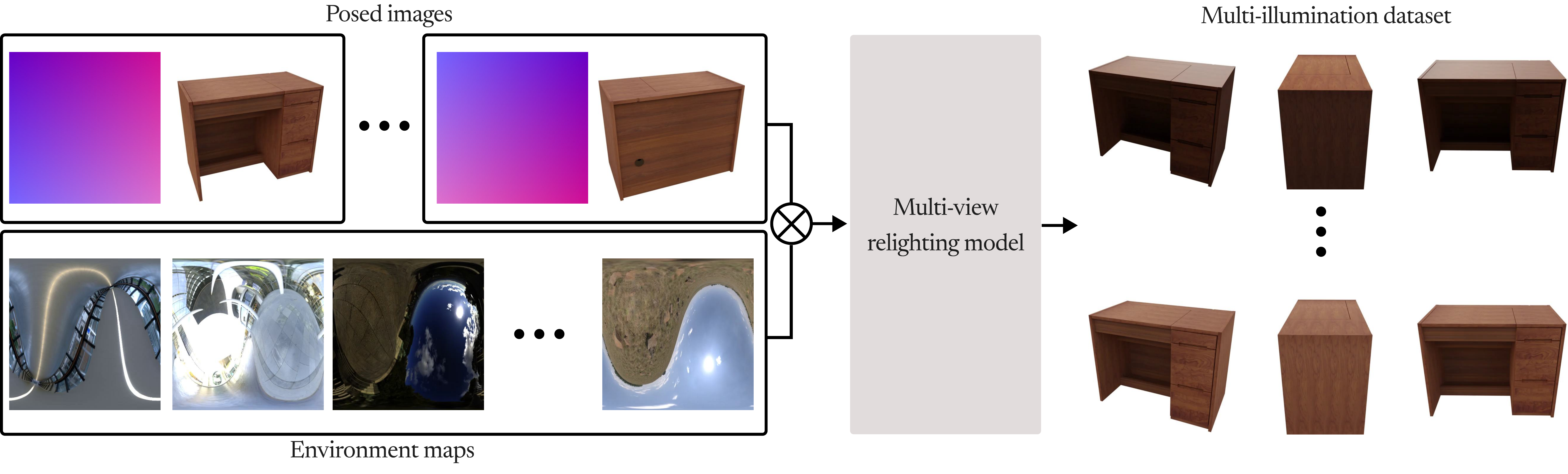}
\caption{\textbf{Multi-view Relight Diffusion}. Our multi-view relighting diffusion model takes as input $N$ posed images illuminated with a consistent, but unknown illumination, represented by camera raymaps and the source pixel values, and an environment map per image that has been rotated to the camera pose. The diffusion model generates images of the same object from the same poses, but lit by an input environment map. To generate our multi-illumination dataset, we repeat this relighting process $M$ times with $M$ environment maps.  
}
\label{fig:overview}
\vspace{-6mm}
\end{figure*}

\section{Method}
\label{SecMethod}
Given a set of $N$ posed images $\mathcal{D}=\{ (I_i, \pi_i) \}_{i=1}^{N}$ of an object, where $I_i$ is the $i$-th image and $\pi_i$ its pose (including camera extrinsic and intrinsic parameters), we are interested in learning the parameters $\theta$ of a relighting function $\mathbf{f}_\theta$, that allows rendering the object from any viewpoint $\pi$ illuminated by any lighting $E$ (e.g.\ an environment map) to produce a new image $I = \mathbf{f}_{\theta}(E, \pi)$. 
In order to learn the parameters $\theta$ of this transformation, we propose to generate a dataset of pairs of posed images with their corresponding illumination to train $\mathbf{f}_\theta$ in a supervised manner. In \refSec{multi_view_diffusion} we will describe how to generate such paired data using a generative relighting diffusion model, and in \refSec{nerf} we explain how to optimize $\mathbf{f}_\theta$, which we implement using a lighting-conditioned, relightable radiance field.

\subsection{Multi-View Diffusion Relighting} \label{sec:multi_view_diffusion}
Our goal is to train a generative multi-view diffusion model $\mathbf{g}$ to relight our given posed images $\mathcal{D}$, which are jointly captured under the same unknown lighting. The diffusion model provides samples from the distribution of possible images $\mathcal{D}_E$ relit by the target illumination $E$:
\begin{equation}
p(\mathcal{D}_{E} | \mathcal{D}, E).
\end{equation}

Our work crucially relies on a multi-view diffusion relighting model to provide consistent relit images, which can then be distilled into a relightable 3D model. This is in contrast with prior work on relighting using diffusion models~\cite{zhao2024illuminerf,jin2024neural_gaffer} which independently relights single images and results in ambiguities that must be resolved at the 3D reconstruction stage.

Our network architecture is inspired by multi-view diffusion architectures such as CAT3D~\cite{gao2024cat3d}, which start with pretrained 2D image diffusion models and inflate them by adding cross-attention layers to process multiple views. We adapt such a scheme for the task of relighting and show our architecture in \refFig{overview}. 
%
Since we use a latent diffusion model (LDM), we first map all images and environment maps into the latent space of the original 2D diffusion model. In particular, the environment map is encoded in a similar manner to Neural Gaffer~\cite{jin2024neural_gaffer}: We use two separate latents corresponding to high dynamic range (HDR) and low dynamic range (LDR). This representation captures both bright details like direct light sources, as well as relatively dim sources like diffuse objects. Like Neural Gaffer, we also use standard tone mapping for the LDR environment map~\cite{hasinoff2016burst,debevec2004high}, and for the HDR encoding we apply logarithmic tone mapping followed by normalization to $[0, 1]$ by subtracting the minimal value and dividing by the maximal value. 
Additionally, for each image $I_i$ in our dataset, we apply a 3D rotation to the environment map to align it with the corresponding camera pose $\pi_i$. 
We combine the HDR and LDR encodings of the environment map with the encoded image and the ray map of the pose of each image by concatenating them, and then apply self-attention. \jiapeng{Details of the network architecture for our multi-view relighting diffusion models are provided in Fig.~1 of the supplementary material.}
\subsection{Relightable Neural Radiance Field} \label{sec:nerf}
Our diffusion model $\mathbf{g}$ generates consistent relit images given a target lighting environment. Our end goal is to obtain a 3D representation of the object that can modify the illumination of the scene \emph{without} additional per-illumination optimization. To do this, we first use the multi-view relighting model to create a new dataset $\mathcal{D}_\text{relit}$ for each object by taking the $N$ images in the original dataset $\mathcal{D}$ and relighting them using a collection of $M$ environment maps $\mathcal{E} = (E_1, ..., E_M)$. The new relit dataset of an object can be written as:
\begin{equation}
\mathcal{D}_\text{relit} = \{\mathbf{g}(I_i, \pi_i, E_j) \colon i=1,...,N, \; j=1,...,M\},
\end{equation}
where $\mathbf{g}(I_i, \pi_i, E_j)$ is the diffusion model's estimate for the $i$th image $I_i$ (whose pose is $\pi_i$) lit by the $j$th environment map $E_j$. $\mathcal{D}_\text{relit}$ contains $N \times M$ images.

\begin{wrapfigure}{r}{0.525\linewidth}
    \centering
    \vspace{-1mm}
    \includegraphics[width=\linewidth]{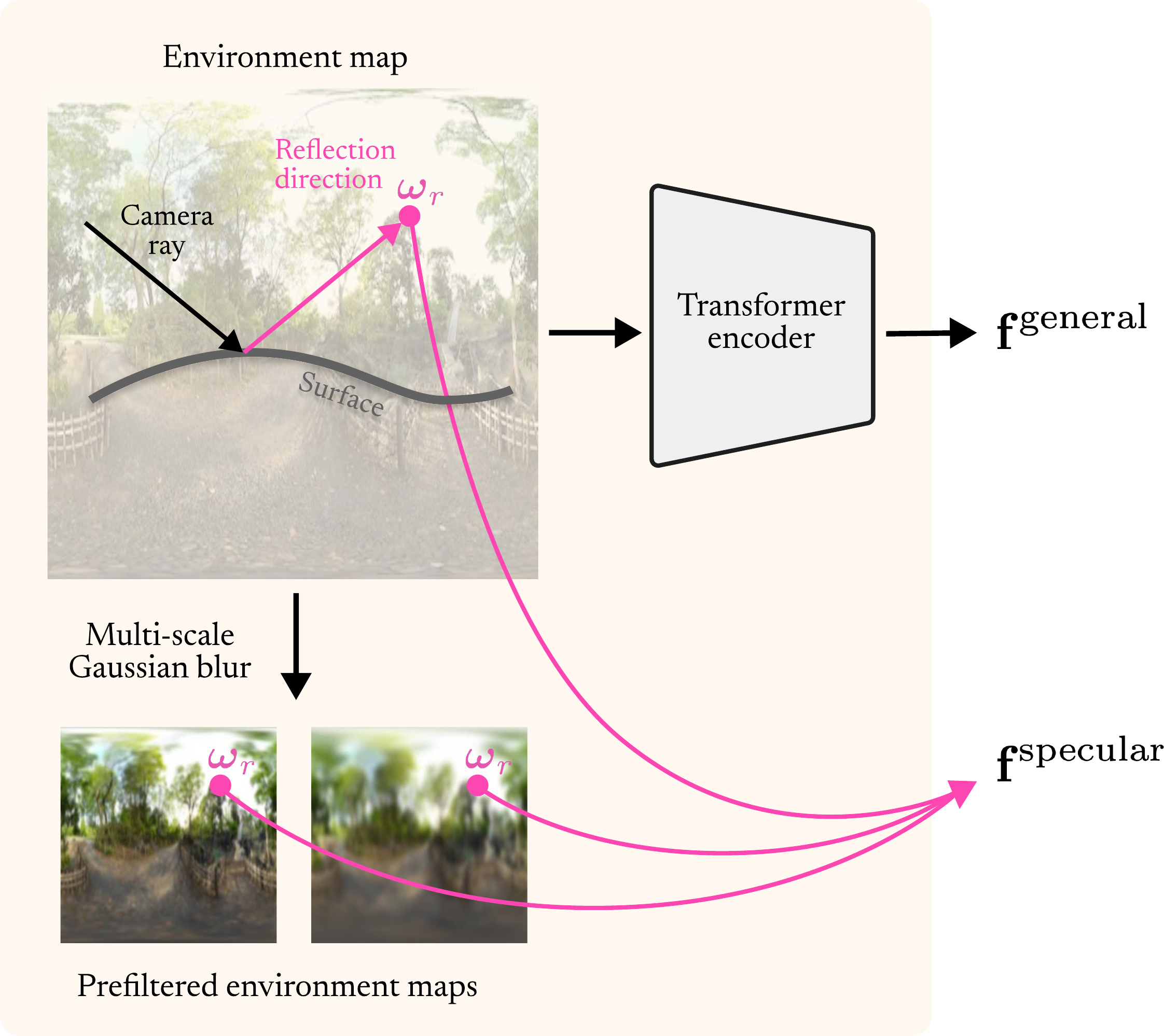}
    \caption{\textbf{Lighting conditioning signals}. We use a combination of two lighting conditioning signals to train the NeRF on our generated multi-illumination dataset. The general lighting encoding $\mathbf{f}^{\text{general}}$ is used for encoding the full environment map in a single embedding, and is obtained using a transformer encoder applied to the entire sphere of incident radiance. The specular encoding $\mathbf{f}^{\text{specular}}$ is composed of the environment map value, as well as prefiltered versions of the environment map, queried at the reflection direction $\boldsymbol{\omega}_r$, which is the direction of the camera ray reflected about the surface normal vector. Combining these two conditioning signals provides the NeRF with all the information necessary for relighting diffuse materials as well as shiny ones, which exhibit strong reflections. }
    \label{fig:conditioning}
    \vspace{-2mm}
\end{wrapfigure}

We then use the multi-illumination dataset $\mathcal{D}_\text{relit}$ to train a NeRF. Since we wish to fit the NeRF model to our dataset with varying lighting so that it generalizes to novel lighting environments \emph{outside} of our training set of illuminants $\mathcal{E}$, care must be taken when designing the model's architecture. We use NeRF-Casting~\citep{verbin2024nerf} as the base NeRF model since it efficiently captures view-dependent appearance, and we modify it to allow for conditioning on the illumination. Crucially, we encode the environment maps using two types of conditioning signals: general conditioning and specular conditioning. The general conditioning encodes the entire environment map into an embedding that is fed to the appearance MLP and is designed as a general-purpose, low-frequency signal for relighting, while the specular conditioning only encodes incident light coming from the specular direction. Its goal is to improve the model's capacity to capture high-frequency reflections (similar to prior work on reconstructing specular reflections in NeRF~\cite{verbin2022refnerf}). The conditioning signals are illustrated in Fig.~\ref{fig:conditioning}.

In order to encode the specular appearance of a camera ray, NeRF-Casting~\cite{verbin2024nerf} casts a small set of reflected rays, traces them through the NeRF's geometry, and volume renders a feature vector $\bar{\mathbf{f}}$ that encodes the scene content observed by these reflected rays. In our setting, we provide the two conditioning signals by concatenating them to the feature vector $\bar{\mathbf{f}}$ of each ray, which is mapped by NeRF-Casting's MLP to the ray's specular color component.

%
\paragraph{General Conditioning.} 
Our \emph{general conditioning} signal encodes the full environment map, and is therefore a complete description of the lighting of the object. To do this, we train an transformer-based encoder which maps the environment map into a vector. While the idea of using a per-lighting encoding is similar to the approach of NeRF-in-the-Wild~\cite{martinbrualla2020nerfw}, a crucial difference is that our embeddings are not optimized to fit to individual images, but they are instead parameterized as a learned mapping from the environment maps. This enables us to render the scene under novel illumination at inference time without requiring any additional training.


To encode the illumination features from the environment map, we utilize a transformer encoder with self-attention. 
Our transformer is trained from scratch, and its architecture is based on ViT-S8 of DINO~\cite{DINO}, followed by a single matrix multiplication $W$ to produce a compact 128-dimensional embedding vector: 
\begin{equation}
    \mathbf{f}^{\text{general}} = W \cdot \text{ViT}(E)
\end{equation}

\paragraph{Specular Conditioning.} 
Although the general conditioning vector theoretically contains all information necessary for relighting, we found it to be insufficient for reconstructing and rendering high-frequency specular highlights. Our \emph{specular conditioning} is designed to fix that by explicitly encoding incident light from the specular reflection direction, similar to the encoding used in prior work for reconstructing reflective objects and scenes~\cite{verbin2022refnerf,verbin2024nerf,liu2023nero}. Instead of only sampling the environment map value at the reflection direction, we use a series of blur kernels centered around the reflection direction to simulate the effects of materials with different roughnesses. The $i$-th component of this conditioning vector can be written as:
\begin{equation} \label{eq:blur}
    \mathbf{f}_i^{\text{specular}} = \int_{\mathbb{S}^2} E(\boldsymbol{\omega}') \text{G}(\boldsymbol{\omega}'; \boldsymbol{\omega}_r, \sigma_i) d\boldsymbol{\omega}'
\end{equation}
where $\text{G}(\boldsymbol{\omega}'; \boldsymbol{\omega}_r, \sigma_i)$ is a Gaussian blur kernel around $\boldsymbol{\omega}_r$ with width $\sigma_i$, and $\boldsymbol{\omega}_r$ is the view direction reflected about the surface normal. For efficiency, we preprocess the environment map by blurring it at all directions using Eq.~\ref{eq:blur}, and then query it at the reflection direction $\boldsymbol{\omega}_r$ during the NeRF optimization stage.




\textbf{Network Predictions.} \jiapeng{Following NeRF-Casting~\cite{verbin2024nerf}, our relightable NeRF takes as input the 3D coordinates, ray direction, general conditioning, and specular conditioning. The geometry MLP predicts density, roughness, normals, and geometry features, while the color MLP outputs the RGB values.}

\begin{figure*}[!hbtp]
    \vspace{-2mm}
    \includegraphics[width=\linewidth]{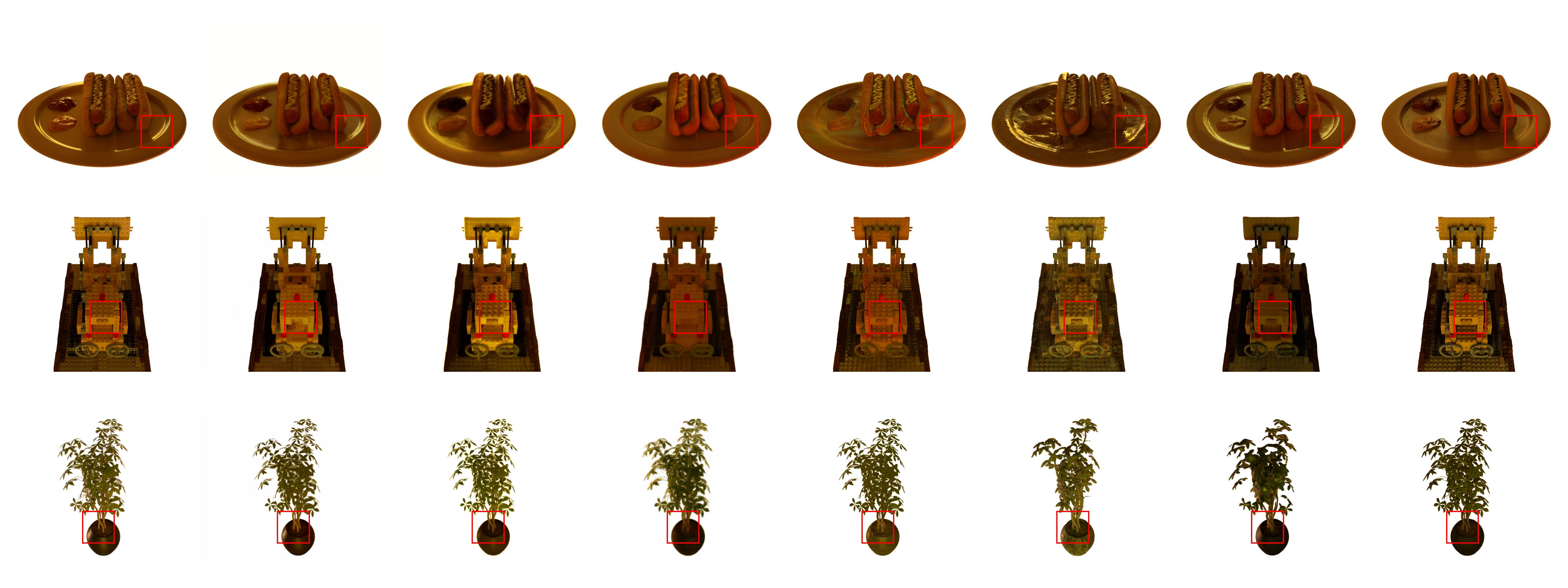}
    \begin{tabularx}{\textwidth}{XXXXXXXX}
        \centering Ground Truth & \centering Ours  & \centering IllumiNeRF & \centering Neural Gaffer & \centering R3DGS 
        & \centering Neural PBIR &\centering NeRO
        & \centering TensorIR
    \end{tabularx}
    \vspace{-2mm}
    \caption{ \textbf{Qualitative comparisons on TensoIR~\cite{jin2023tensoir}.} All renderings are rescaled to the image resolution of the ground truth. Compared to previous works, our method recovers more plausible specular highlights and more accurate colors as indicated with the red box. }  
    \label{fig:tensoir}
    \vspace{-4mm}
\end{figure*}

\section{Experimental Setup}
\label{SecExper}
\subsection{Datasets}
\paragraph{Training datasets.} To train multi-view relighting diffusion, we use a dataset of 400k synthetic 3D objects, including 100k from Objaverse~\cite{deitke2023objaverse}. 
Each object is rendered in 64 views $\times$ 16 HDR illuminations, with environments sampled from Polyhaven~\cite{zaal2021polyhaven} (590 maps) and augmented via random up-axis rotations.

\paragraph{Evaluation datasets.}
For relighting evaluations, we used two datasets: TensoIR~\cite{jin2023tensoir} and Stanford-ORB~\cite{kuang2024stanford}. TensoIR is a synthetic benchmark, which contains renderings of four objects under six lighting conditions. 
We use the train split of 100 views with ``sunset'' lighting condition as inputs for relightable NeRF. We then evaluate 200 novel views under other five environment maps, including ``bridge'', ``fireplace'', ``forest'', ``city'', and ``night''. 
In total, we have 4,000 renderings for evaluation metric calculation. 
Stanford-ORB is real-world benchmarks by data capture in the wild. It has 14 objects composed of various materials.
Each object is captured under three distinct lighting conditions, producing a total of 42 (object, lighting) combinations. 
Following its evaluation protocol, we use images of an object under a single lighting condition and evaluate novel views under the two target lighting settings.
%

\subsection{Implementation details}
\paragraph{Multi-View Diffusion model.} 
We fine-tune our model starting from a pre-trained latent image generation model, as described in~\cite{rombach2022high}. The multi-view denoiser is derived from the CAT3D network architecture~\cite{gao2024cat3d}, with modifications to the input channel dimensions to align with our relighting task. The inputs to our model are images with resolution $512\times 512$ which are encoded to $64\times 64\times 8$ latents. We chose the number of views to be 64. The model was trained on 128 TPU v5 chips using a learning rate of $10^{-4}$, with a total batch size of 128 for 360k iterations. After training, we generate the multi-illumination dataset by running our relighting diffusion inference on 111 environment maps.

\paragraph{Relightable NeRF model.}
We train our NeRF on 8 H100s for 500k steps. We use a $512\times 512$ resolution environment map as the target illumination. We sample each reflection rays 3 times; once using a point sample on the full resolution environment map, and then using Gaussian kernels of sizes $20\times 20$ and $40\times 40$ pixels in radius with $\sigma_i$ values of 10 and 20 respectively (see Fig.~\ref{fig:conditioning}). In order to maximize the number of Illumination conditions we use for training, we make several reductions to the size of model relative to the NeRF-Casting architecture. We lower the batch size to 1,000 and increase the number of training steps to 500,000. We also decrease the size of the ``bottleneck'' vector $\mathbf{b}$ in both the geometry and appearance MLPs relative to NeRF-Casting. 
Please refer to supplementary material for more details on the base architecture.

\subsection{Baselines}
We compare our method against existing inverse rendering methods including NeRFactor~\cite{zhang2021nerfactor}, InvRender~\cite{zhang2022modeling}, PhySG~\cite{zhang2021physg}, NeRD~\cite{boss2021nerd}, NVDiffRecMC~\cite{hasselgren2022nvdiffrecmc}, NVDiffRec~\cite{NVDiffRec}, Neural-PBIR~\cite{Neural-PBIR}, NeRO~\cite{liu2023nero},TensoIR~\cite{jin2023tensoir}, and recent single-view relighting diffusion methods IllumiNeRF~\cite{zhao2024illuminerf} and Neural Gaffer~\cite{jin2024neural_gaffer}. We also include the most recent Gaussian splatting-based inverse rendering method R3DG~\cite{R3DG}.

\subsection{Evaluation Metrics}
We evaluate the relighting rendering quality by PSNR, SSIM~\cite{SSIM}, and LPIPS-VGG~\cite{LPIPS}. 

\begin{wraptable}{R}{8cm}
    \vspace{-6mm}
	\begin{center}
        \footnotesize
		\begin{tabular}{*{4}{c}}
                \toprule
			    Method  & PSNR $\uparrow$ & SSIM $\uparrow$  & LPIPS $\downarrow$   
                     \\ 
                \midrule
                NeRFactor~\cite{zhang2021nerfactor} &  23.38  &  0.908  &  0.131  \\ 
                InvRender~\cite{zhang2022modeling} &  23.97  &  0.901 &  0.101 \\
                TensoIR~\cite{jin2023tensoir} &  28.58  &  0.944 &  0.081 \\ 
                
                Neural-PBIR~\cite{Neural-PBIR} & 27.09 & 0.925 & 0.085 \\

                NeRO~\cite{liu2023nero} & 27.00 & 0.935	& 0.074 \\
                
                %
                R3DG~\cite{R3DG}  & 29.05   & 0.937  & 0.080 \\
                NeuralGaffer~\cite{jin2024neural_gaffer}  & 27.30   & 0.918  & 0.122 \\
                IllumiNeRF~\cite{zhao2024illuminerf}  &  \tabsecond{29.71}  &  \tabsecond{0.947}  & \tabsecond{0.072} \\
                \midrule
                Ours  &  \tabfirst{30.74}  & \tabfirst{0.950}  & \tabfirst{0.069} \\
                \bottomrule
        \end{tabular}
        \caption{\textbf{TensoIR benchmark~\cite{jin2023tensoir}.} We evaluate all four objects in the benchmark, each under five novel lightings. Each object is rendered from 200 views for novel view evaluation under each lighting. Thus, we have 4,000 renderings in total for quantitative evaluation. 
        \tabfirst{Best} and \tabsecond{2nd-best} are highlighted.  
        }
        \label{tab:tensoir}
        \end{center}
        \vspace{-10mm}
\end{wraptable}
\section{Results}
Our method is the top-performing technique on existing relighting benchmarks for synthetic and real-world objects. Furthermore, it can render images from novel viewpoints under novel illuminations at interactive speeds (0.384 seconds per frame). The only other methods that achieves similar relighting and rendering speeds are ones based on Gaussian splatting combined with inverse rendering, such as R3DG (0.415 seconds per frame), but the quality of these methods is significantly lower than of generative methods or inverse rendering methods that are based on NeRF.

\subsection{TensoIR benchmark}
In Fig.~\ref{fig:tensoir},
our method achieves state-of-the-art performance on the TensoIR benchmark, improving upon the ability of prior works to capture specularities in the reflective ``hot dog'' and ``ficus'' scenes while accurately capturing diffuse appearance from global illumination in the ``lego''  and ``armadillo''  scenes. 
The superiority of our method can also be verified by quantitative results in Tab.~\ref{tab:tensoir}.

\begin{table}[t]
        \vspace{-8mm}
    \setlength{\tabcolsep}{15.5pt} 
	\begin{center}
        \footnotesize
		\begin{tabular}{*{5}{c}}
                \toprule
			    Method  & PSNR-H $\uparrow$ & PSNR-L $\uparrow$
                & SSIM $\uparrow$  & LPIPS $\downarrow$  
                     \\ 
                \midrule
                NVDiffRecMC~\cite{NVDiffRecMC} $\dagger$ & 25.08  &  32.28  &  0.974  & \tabsecond{0.027} \\
                NVDiffRec~\cite{NVDiffRec} $\dagger$ & 24.93  & 32.42  & 0.975   & \tabsecond{0.027} \\
                \midrule
                PhySG~\cite{zhang2021physg} &  21.81  &   28.11   &  0.960   &  0.055 \\
                NVDiffRec~\cite{NVDiffRec} &   22.91 & 29.72 & 0.963  & 0.039\\ 
                NeRD~\cite{boss2021nerd}  &  23.29 & 29.65  & 0.957 & 0.059 \\
                NeRFactor~\cite{zhang2021nerfactor}  & 23.54 & 30.38 & 0.969 & 0.048 \\
                InvRender~\cite{zhang2022modeling}  & 23.76 & 30.83 & 0.970 & 0.046 \\
                NVDiffRecMC~\cite{NVDiffRecMC}  &  24.43 & 31.60 & 0.972 & 0.036 \\
                Neural-PBIR~\cite{Neural-PBIR}  &  \tabsecond{26.01} & \tabfirst{33.26} & \tabsecond{0.979} & \tabfirst{0.023} \\
                R3DG~\cite{R3DG}  & 21.25   &  27.50 &  0.962  & 0.063   \\
                Neural Gaffer~\cite{jin2024neural_gaffer}  & 23.16   &  29.94  &  0.966   &  0.047  \\ 
            IllumiNeRF~\cite{zhao2024illuminerf}  & 25.56 & 32.74 & 0.976 & \tabsecond{0.027} \\
                \midrule
                 Ours  &  \tabfirst{26.21}  &  \tabsecond{32.91}  &  \tabfirst{0.980}  &   \tabsecond{0.027}  \\
                \bottomrule
        \end{tabular}
        \vspace{2mm}
        \caption{\textbf{StandfordORB benchmark~\cite{kuang2024stanford}.} 
        We evaluate fourteen objects captured in the real world. Each object was captured in three different lightings. For each object-lighting pair, we evaluate novel view renderings under the other two lighting. The benchmark contains 840 renderings in total.
        $\dagger$ denotes models trained with the ground-truth 3D scans and pseudo materials optimized from light-box captures. \tabfirst{Best} and \tabsecond{2nd-best} are highlighted.
        }
        \label{tab:standfordorb}
        \end{center}
        \vspace{-8mm}
\end{table}
\begin{figure*}[t]
    \vspace{-8mm}
    \begin{adjustbox}{angle=90}
    \begin{tabular*}{0.01\linewidth} {@{\extracolsep{\fill}}p{0.08\linewidth} p{0.12\linewidth}p{0.08\linewidth}p{0.08\linewidth}p{0.08\linewidth}p{0.08\linewidth}p{0.10\linewidth}p{0.08\linewidth} }
        \centering\small InvRender & \centering\small NVDiffRecMC  &  \centering\small Neural-PBIR & \centering\small R3DGS  &  \centering\small Neural Gaffer
        & \centering\small IllumiNeRF & \centering\small Ours & \centering\small GT  \\
        \end{tabular*}
    \end{adjustbox}
    \hspace{-4mm}
        \includegraphics[width=.95\linewidth]{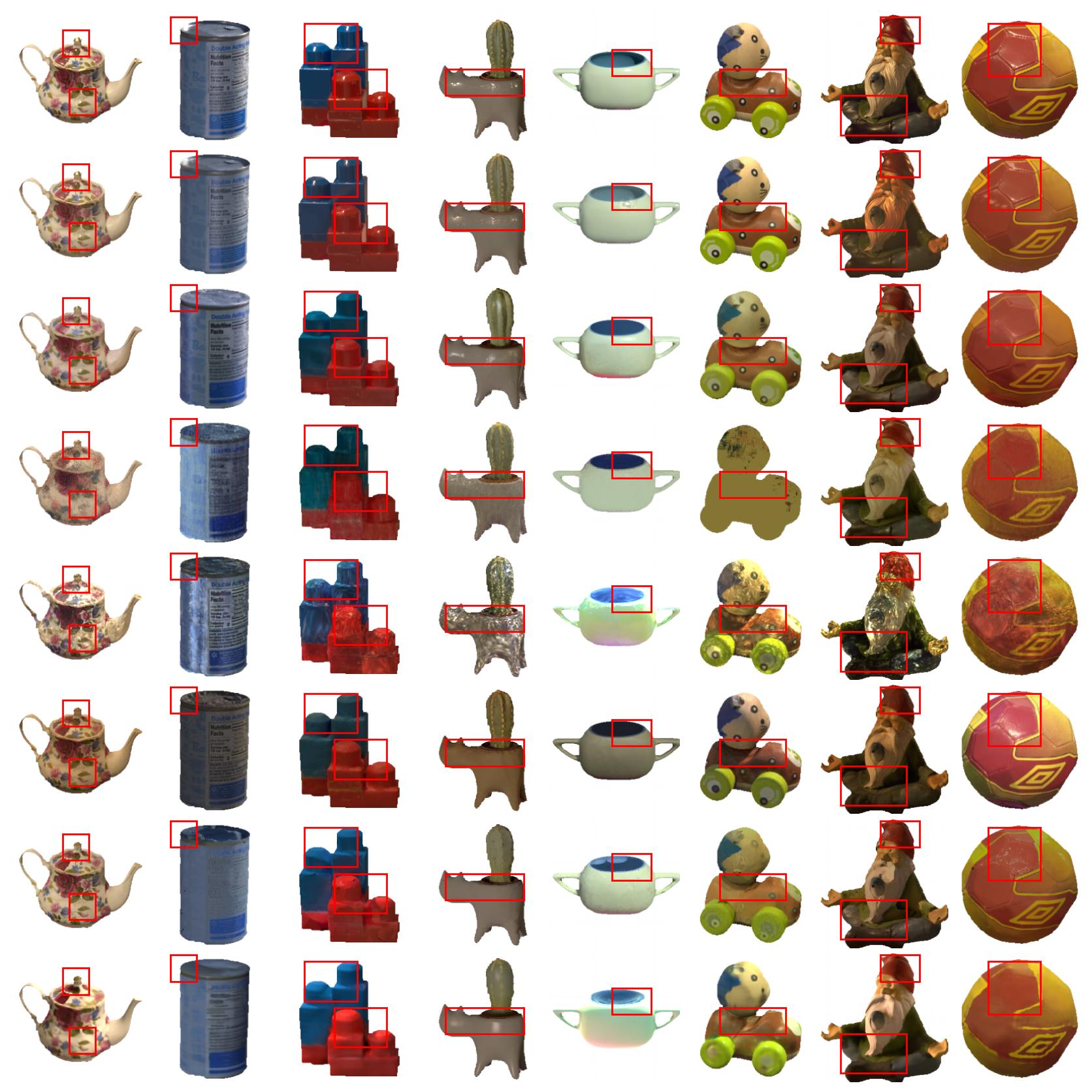}
        \caption{ \textbf{Qualitative comparisons on Stanford-ORB~\cite{kuang2024stanford}.} Renderings from all methods are rescaled to the image resolution of the ground truth.  Compared to previous work, our method produces high-fidelity renderings with more faithful specular reflections highlighted in the red boxes. 
        }
        \label{fig:standfordorb}
\end{figure*}
\subsection{Stanford-ORB benchmark}
In Fig.~\ref{fig:standfordorb}, our method also achieves state-of-the-art performance on the Stanford-ORB dataset, and is most effective at reflective objects like the ``baking'' and ``ball'' scenes where consistent reflections clearly show up in the reconstruction. We provide quantitative comparisons in Tab.~\ref{tab:standfordorb}, where our method outperforms others in PSNR-H and SSIM, and acheive second-best results in PSNR-L and LPIPS. As discussed in IllumiNeRF~\cite{zhao2024illuminerf}, our results are also qualitatively superior to those of Neural-PBIR~\cite{Neural-PBIR}, but they are worse quantitatively due to the mostly-diffuse renderings of Neural-PBIR.

\subsection{Real-world Objects} 

Finally, we demonstrate the ability of our method to relight ``in-the-wild'' captures of real objects with spatially-varying material properties under natural lighting in Fig.~\ref{fig:Teaser}.
Since we have no ground truth images for real-world relighting evaluation, we only show qualitative results.
%
%
Our method captures convincing specularities on the wood and metal parts of the model sewing machine, as well as accurate shadows cast by the arm of the sewing machine. 

\subsection{Ablation Studies}

\begin{table}[t]
        \vspace{-3mm}
    \setlength{\tabcolsep}{15.5pt}
	\begin{center}
        \footnotesize
		\begin{tabular}{*{6}{c}}
                \toprule
			    Method   & PSNR $\uparrow$ & SSIM $\uparrow$  & LPIPS $\downarrow$   &  \\ 
                \midrule
               (a) No blurring in specular conditioning  &  \tabsecond{31.35}  & \tabsecond{0.90} & 0.080 \\
               (b) No specular conditioning  &  30.00  & 0.88  & \tabsecond{0.079} \\
                (c) No general conditioning       &  21.59  & 0.70 & 0.130  \\
                (d) Per-image appearance embeddings   &  19.12  & 0.62  & 0.160 \\
                (e) $128\times 128$ environment map  &  29.26  & 0.89 & 0.082 \\
                (f) $64 \times 64$ environment map  &  27.69  & 0.86 & 0.110 \\
                Our full model, 64-view, 111 envmaps & \tabfirst{31.88}  & \tabfirst{0.91} & \tabfirst{0.075} \\
                \bottomrule
        \end{tabular}
        \vspace{1mm}
        \caption{\textbf{ Ablation studies on the ``hotdog'' scene from TensoIR~\cite{jin2023tensoir}.} See the text and Fig.~\ref{fig:ablation} for corresponding qualitative results and additional explanations. \tabfirst{Best} and \tabsecond{2nd-best} are highlighted.
        }
        \label{tab:ablation}
        \end{center}
       \vspace{-10mm}
\end{table}
\begin{figure*}[!htbp]
    \vspace{-4mm}
    \includegraphics[width=\linewidth, trim=0 50 0 0, clip]{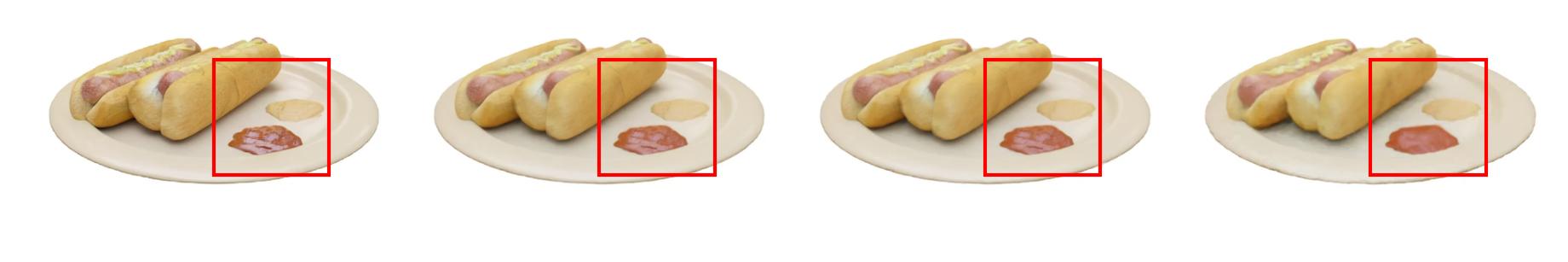}
    \begin{tabularx}{\textwidth}{XXXX}
        \centering \small Ground truth & \centering \small Our model  & \centering \small (a) & \centering \small (b) 
    \end{tabularx}
    \includegraphics[width=\linewidth, trim=0 50 0 0, clip]{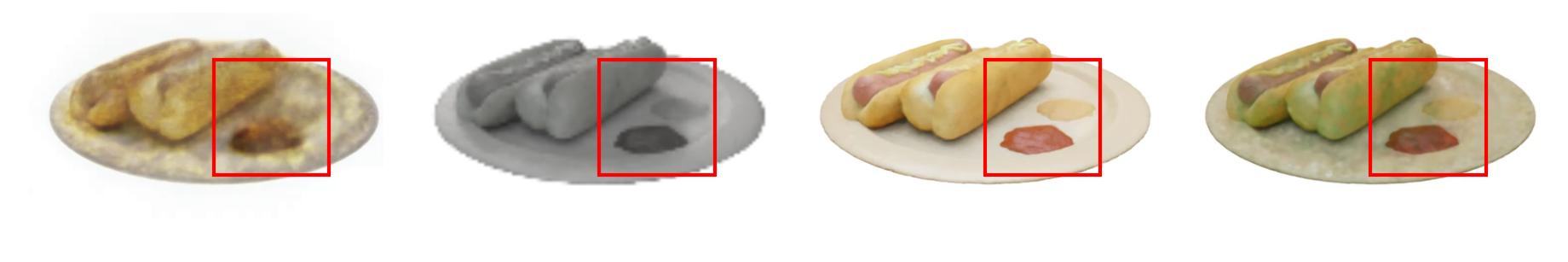}
    \begin{tabularx}{\textwidth}{XXXX}
    \centering \small (c) & \centering \small (d) & \centering \small (e) & \centering \small (f)
    \end{tabularx}
    \vspace{-2mm}
    \caption{ \textbf{Qualitative ablation studies.} We compare a ground truth image and the relighting results of our model with a set of ablations (a)-(f). (a) Using specular conditioning without blurring results in high-quality images with slightly lower metrics, whereas (b) removing the specular conditioning altogether greatly reduces the accuracy of specular highlights. (c) We observe that our model produces significant artifacts when the general conditioning is removed or (d) replaced with per-image appearance embeddings; and that providing environment maps at lower resolutions of (e) $128\times 128$ or (f) $64\times 64$ tends to blur specular highlights and introduce rendering artifacts.
    }
    \label{fig:ablation}
     \vspace{-3mm}
\end{figure*}
%
We next perform an ablation study of the different components of our method, done on the ``hotdog'' scene of the TensoIR benchmark, chosen since it has the most interesting materials in that dataset. Each ablation is reported in a row of Tab.~\ref{tab:ablation} and its corresponding column in Fig.~\ref{fig:ablation}. 

\paragraph{Multi-scale specular conditioning.}
Our multi-scale specular conditioning features, which are computed by blurring the environment map using multiple kernel sizes as in Eq.~\ref{eq:blur}, are provided to the model. Each scale is designed to approximate the incident light averaged over a set of directions corresponding to a particular surface roughness. When we skip this blurring operation and use the environment map values directly, our model can still represent highly specular regions (see Fig.~\ref{fig:ablation}(a)), but struggles more with rough or diffuse surfaces, leading to a drop in reconstruction metrics.

\paragraph{Specular conditioning.}
Next we remove the specular conditioning signal altogether, which leads to another small drop in reconstruction metrics as well as a qualitative drop in the accuracy of rendered specular highlights.

\paragraph{General conditioning.}
Removing the general conditioning signal from our network results in significant artifacts and poor reconstruction metrics, since the general conditioning is the main mechanism for providing the target illumination to the relighting model.

\paragraph{Per-image appearance embeddings.}
An alternative to our conditioning signals is a per-image appearance embedding vector, similar to the GLO codes in NeRF-in-the-Wild~\cite{martinbrualla2020nerfw}. While this allows the model to be trained on multiple illuminations, it does not generalize to new unseen lights. Additionally, unlike our model, we found that the embedding vectors do not scale well to a large number of illumination conditions, resulting in significantly worse qualitative and quantitative results.



\paragraph{Environment map resolution.}
Our full model uses conditioning signals based on an environment map of resolution $512 \times 512$. Computing the conditioning signals from environment maps of size $128\times 128$ results in loss of detail in the specular highlights. Further lowering the resolution to $64\times 64$ results in rendering artifacts even for diffuse surfaces.

\paragraph{Number of views.}  
Our full model learns the joint distribution of 64-view relighting. To analyze the impact of the number of views in the diffusion model, we compare relighting novel view synthesis results using 4, 16, and 64-view diffusion on the hotdog scene of TensoIR dataset. As shown in Tab.~\ref{tab:ablation_supple}, the 64-view diffusion model consistently outperforms the others across all metrics. This indicates that jointly denoising more views leads to more consistent and higher-quality relit images.

\begin{table}[t]
    \setlength{\tabcolsep}{15.5pt}
	\begin{center}
        \footnotesize
		\begin{tabular}{*{6}{c}}
                \toprule
			    Method   & PSNR $\uparrow$ & SSIM $\uparrow$  & LPIPS $\downarrow$   &  \\ 
                \midrule
                Ours full model, 4 view, 111 envmaps  &  	31.04 	& 0.86 &	0.082 \\
                Ours full model, 16 view, 111 envmaps  &  31.86	& 0.90	& 0.077 \\
                Our full model, 64-view, 10 envmaps  &  	29.12 &	0.82 &	0.086 \\
                Our full model, 64-view, 50 envmaps  &  31.78  &	0.88  &	0.079 \\
                 Our full model, 64-view, 111 envmaps & \tabsecond{31.88}  & \tabsecond{0.91} & \tabsecond{0.075} \\
                Our full model, 64-view, 150 envmaps  & \tabfirst{31.90}  & \tabfirst{0.92} & \tabfirst{0.075} \\
                \bottomrule
        \end{tabular}
        \vspace{1mm}
        \caption{\textbf{ Ablation studies of number of views and illulimantions on the ``hotdog'' scene from TensoIR~\cite{jin2023tensoir}.}\tabfirst{Best} and \tabsecond{2nd-best} are highlighted.
        }
        \label{tab:ablation_supple}
        \end{center}
        \vspace{-2mm}
\end{table}
\paragraph{Number of environment maps.}
\jiapeng{
We also study the effect of environment map count on relightable NeRF training by using 10, 50, 111, and 150 illuminations. As shown in Tab.~\ref{tab:ablation_supple}, more illuminations generally improve relighting performance, which saturates around 111 maps.
}
%

\begin{figure}[t]
    \vspace{-10mm}
    \centering
    \includegraphics[width=\linewidth]{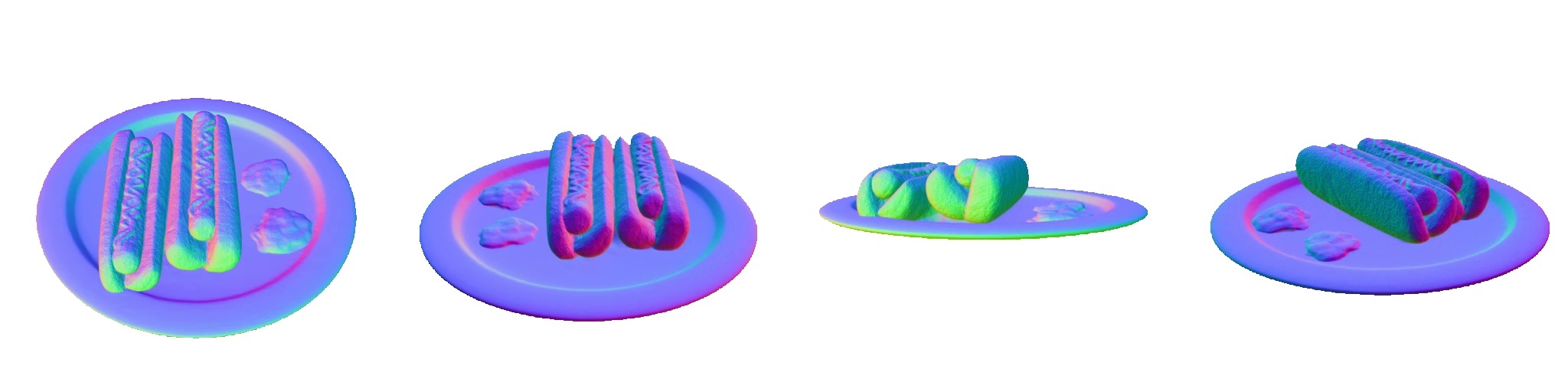} 
    \includegraphics[width=\linewidth]{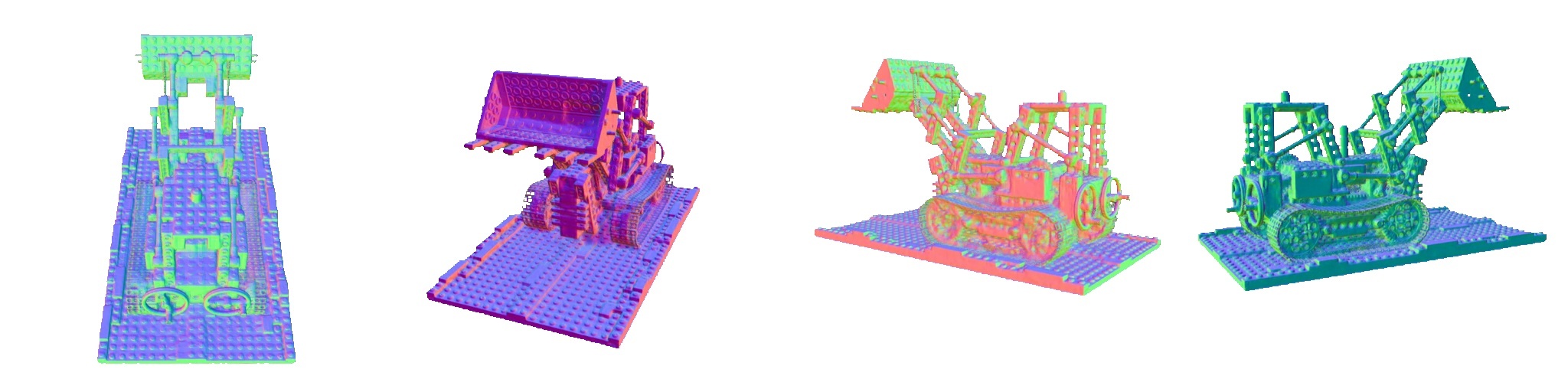} 
    \vspace{-6mm}
    \caption{\textbf{Normal map rendering of our relightable NeRF on objects from the TensoIR dataset. }}
    \label{fig:normal}
    \vspace{-4mm}
\end{figure}

\paragraph{Normal map visualization.} \jiapeng{ As shown in Fig.~\ref{fig:normal}, our relightable neural radiance fields can render high-quality normal maps, which are comparable to prior inverse rendering techniques like TensoIR, as well as novel view synthesis approaches like NeRF-Casting which explicitly encourage geometry to be more surface-like. 
}

 \subsection{Limitations}
While our method improves 3D object relighting by achieving more accurate specularities and fast inference, it can still be further improved in several aspects.
First, although our relighting diffusion model is trained on objects with materials varying in diffuse albedo and roughness, which are the main sources of variation in real-life materials, we did not train on objects exhibiting phenomena such as subsurface scattering, refraction, or volumetric effects. Expanding our training data to include these complex materials would improve robustness and generalizability to real data.
Second, we use environment maps to define lighting conditions, which assumes that the light sources are infinitely far away from the object. Exploring more general lighting models containing near-field illumination components could enhance realism in diverse illumination scenarios.
Finally, our approach focuses on object-centric scenes, and extending it to large-scale scene relighting would be an exciting direction for future research.

\section{Conclusion}
\label{SecCon}
This paper introduces a novel method for 3D object relighting, enabling fast, feed-forward relighting during inference. By modeling a virtual light stage with a generative multi-diffusion model, we create a diverse dataset of multi-illumination images. This dataset serves as a prior to train a light-conditioned Neural Radiance Field (NeRF) model, which subsequently learns to render the object under arbitrary target illumination conditions. Existing methods for 3D relighting often rely on inverse rendering techniques, constrained by shading models limited to specific material types, or they directly generate relit NeRFs, embedding the lighting within the model itself. This requires retraining the NeRF for each new lighting condition. In contrast, our proposed model exhibits generalization across diverse lighting conditions at inference, facilitating efficient feed-forward relighting. Experimental results demonstrate the effectiveness of our method in relighting complex real-world objects with high fidelity.

\paragraph{Acknowledgement.}
 We would like to thank Xiaoming Zhao, Rundi Wu, Songyou Peng, Ruiqi Gao, Stan Szymanowicez, Hadi Alzayer, Ben Poole, Aleksander Holynski, Jason Zhang, Jonathan T. Barron, Peter Hedman, Alex Trevithick, and Jiahui Lei 
for their valuable contributions. We also extend our gratitude to Shlomi Fruchter, Kevin Murphy, Mohammad Babaeizadeh, Han Zhang and Amir Hertz for training the base text-to-image latent diffusion model.

{
    \small
    \bibliography{neurips_2025}
}

\newpage
\appendix
\section*{Appendix -- ROGR: Relightable 3D Objects 
    using Generative Relighting}
\section{Supplementary Webpage and Videos}
 We suggest readers check the website in our supplementary material. We provide more video renderings, including ablation studies and result comparisons on TensoIR~\cite{tensoir}, StanfordORB~\cite{kuang2024stanford}, and real-world objects.

\section{Additional Implementation Details}

\subsection{Multi-view Relighting Diffusion}
We implement our multi-view relighting diffusion model using JAX~\cite{bradbury2018jax}. It is initialized from a pre-trained latent diffusion model for text-to-image generation, similar to StableDiffusion~\cite{rombach2022high}. Our model denoises multiple noisy latents of size $64 \times 64  \times 8$ and decodes them into multi-view relit images of size $512 \times 512 \times 3$. Since our model is not conditioned on text prompt, we only feed empty strings to the CLIP text encoder~\cite{radford2021learning}.

\begin{figure}[!htbp]
    \centering
    \includegraphics[width=.85\linewidth]{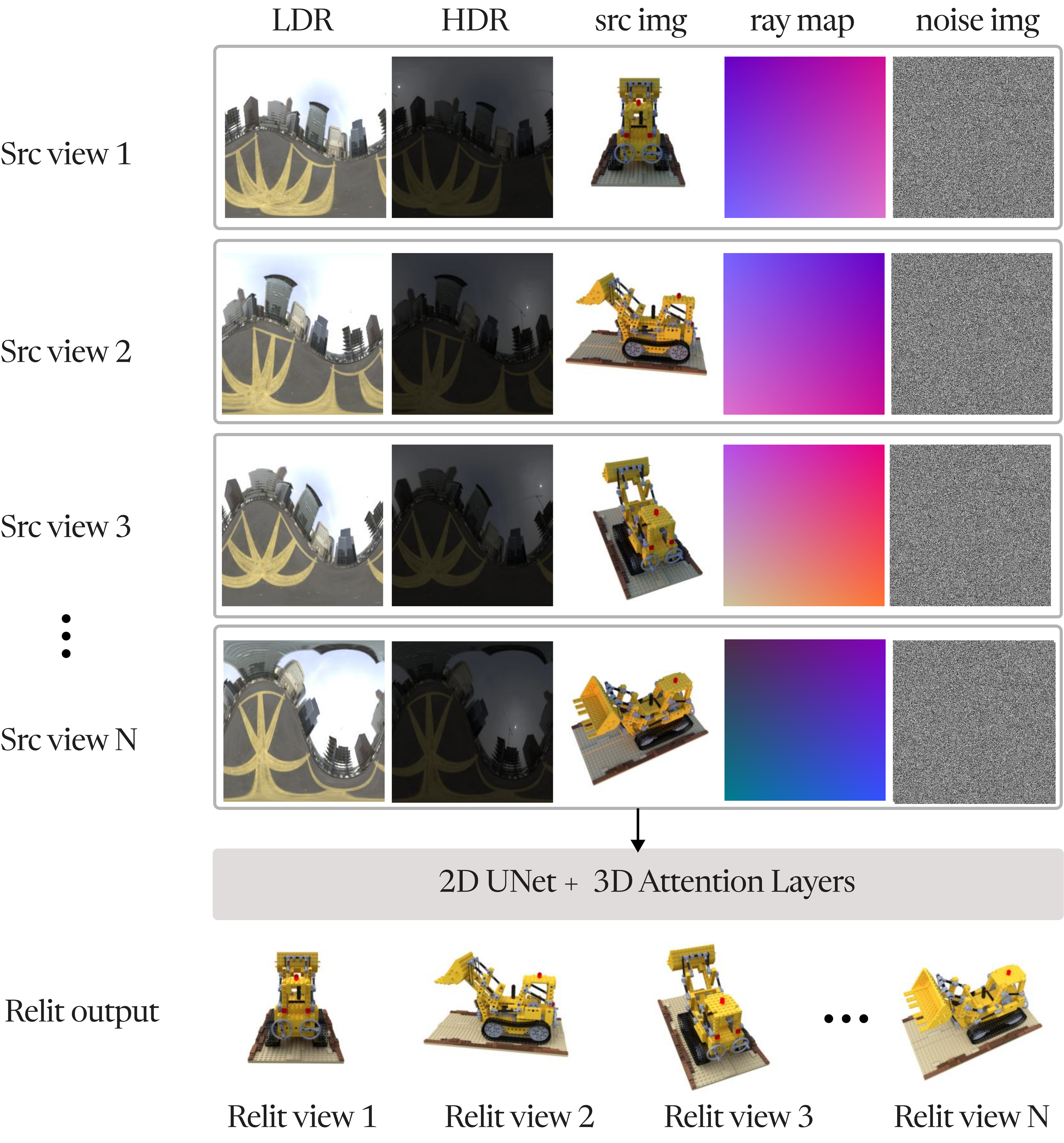}
    \caption{\textbf{Multi-view Relighiting Diffusion Models.} For each view, we concatenate noisy image latents, raymaps containing pose information, source image latents, HDR and LDR environment latents as inputs, and feed them into a multi-view denoiser network that is implemented by 2D UNet with additional 3D Attention layers. 2D UNet individually processes the latent feature from each view. 3D Attention layers flatten multi-view image latents into a 1D sequence, and then perform self-attention to exchange information across different pixels and views. }
    \label{fig:mvrelit}
\end{figure}

In Fig.~\ref{fig:mvrelit}, we provide the detailed inputs of our relighting diffusion model. While our diffusion model receives and produces image latents, we depict them as images for better clarity.  We encode RGB images under source lighting, HDR, and LDR target environment maps into latents of size $64 \times 64 \times 8$. The raymaps consist of ray origins and ray directions corresponding to image pixels.  We downscale them from the original image resolution $512 \times 512 \times 6$ to the latent size $64 \times 64 \times 6$. We concatenate the source image latent, HDR, LDR environment latents, raymaps, and noisy latents along the channel dimension to form a new feature of size $64 \times 64 \times 38$, and then feed it into a multi-view diffusion network to produce clean target latents of size $64 \times 64 \times 8$.  Our denoiser network is based on CAT3D~\cite{gao2024cat3d}.  Please refer to Fig. 7 of CAT3D for the network architecture details.

During training, we use the DDPM schedule, with beta values that linearly increase from $8.5 \times 10^{-4}$ to $1.2 \times 10^{-2}$ over 1024 steps. We use noise prediction as our diffusion objective.  The model was trained on 128 TPU v5 chips using a learning rate of $10^{-4}$, with a total batch size of 128 for 360k iterations and 10K warm-up steps. We adopted a progressive training scheme, where we first trained a 4-view diffusion model for 300k steps, and then fine-tuned it for 16-view diffusion for 15k steps, and finally fine-tuned it for 64-view diffusion for 45 steps. We keep the learning rate as $10^{-4}$ when we fine-tune the model to relight the large number of views. We enable classifier-free guidance (CFG)~\cite{ho2022classifier} by randomly dropping the HDR and LDR environment maps with a probability of 0.1.
During inference, we use the DDIM schedule~\cite{song2020denoising} with 50 sampling steps and the classifier-free weight is set to 3.0.

\subsection{Relightable NeRF}
Our relightable NeRF is mainly based on NeRF-Casting~\cite{verbin2024nerf}, which implicitly learnt the accumulated reflection features to model accurate and detailed reflections. Instead, we choose to explicitly learn these from given enviroment maps. Given a single ray with origin $\vec{o}$ and direction $\vec{d}$, we sample $N$ points $\vec{x}^{(i)}$ along the ray and use multi-resolution hash grid and MLP to encode $\vec{x}^{(i)}$ into density $\vec{\tau}^{(i)}$, roughness $\vec{\rho}^{(i)}$, and surface normal $\vec{n}^{(i)}$. Then, they are alpha composited to compute a single expected termination point $\vec{\overline{x}}$, a von Mishes-Fisher distribution (vMF) width $\overline{k}$, and surface normal $\vec{\overline{n}}$. Next, we construct a reflection cone by reflecting $\vec{d}$ around the micro-surface to obtain a vMF distribution over reflected rays $vMF(\vec{d}', \overline{k})$. We sample $K=5$ reflected rays with location $\vec{o}'$ and $\vec{d}'_j$. We cast these rays and sample $N'$ points $\vec{x}^{(i)}_j$ along each reflection ray. $N'$ points $\vec{x}^{(i)}_j$ are encoded into $N'$ densities.  Based on location $\vec{o}'$ and $\vec{d}'_j$, we can query illumination information from environment maps and encode them with $\vec{x}^{(i)}_j$ into features $\vec{f}^{(i)}_j$ through an MLP.
These features  $\vec{f}^{(i)}_j$ are alpha-composed along each ray to get per-ray reflection feature ${\vec{\overline{f}}^{(i)}}_j$, which can be further averaged into a single reflection feature $\vec{f}$. Finally, $\vec{f}$, bottleneck geometry feature $\vec{b}^{(i)}$, mixing coefficients $\beta^{(i)}$ and view direction $\vec{d}$ are feed into color decoder to predict the color value of $\vec{x}^{(i)}$. 

\section{Data}
\subsection{Training data preprocessing}
We use Objaverse~\cite{deitke2023objaverse} and an internal dataset containing high-quality 3D assets as our training data. The Objaverse dataset is released under the Apache-2.0 license.

To ensure high-quality renderings for training, we filter out low-quality 3D assets from Objaverse using the object list provided in~\cite{tang2024lgm}, and further exclude (semi-)transparent objects, as our focus is on reflective and shiny surfaces. For 3D assets lacking texture or material information, we assign a uniform color sampled from $[0, 1]$ as texture. We randomly sample three values from $[0, 1]$ as the diffuse, roughness, and metallic terms of the material model.

Our relighting diffusion model requires multi-view images under diverse lighting conditions. To this end, we use 590 equirectangular environment maps from~\cite{zaal2021polyhaven}. For each object, we randomly sample 64 camera poses on a sphere centered around the object. The camera distance ranges from $[0.5, 2.0]$. For each camera view, we randomly select 16 environment maps, augmenting each with a random horizontal shift. Then we 
 use Blender's Cycles path tracer to render images at a resolution of 512$\times$512, with 512 samples per pixel. 

\subsection{Evaluation benchmarks.}
We use TensoIR~\cite{tensoir} and StanfordORB~\cite{kuang2024stanford} datasets for relighting evaluations. TensoIR is under the MIT license. StanfordORB does not specify a license in the official GitHub repository.

%

\subsection{Real-world data capture}
We capture real-world objects to evaluate our relighting model using a Sony DSLR camera.  Each object is placed on a table, and a handheld camera is used to record an image sequence by moving around the object. Each sequence contains approximately 160 to 300 frames. Camera intrinsics and extrinsics are estimated using COLMAP\cite{schonberger2016structure} for NeRF training. For foreground segmentation, we apply the pre-trained SAM~\cite{kirillov2023segment} model. We will release our captured dataset upon publication under a non-commercial academic license. 
%
%

\subsection{Evaluation metrics}
We evaluate the relighting rendering quality by PSNR, SSIM~\cite{SSIM}, and LPIPS-VGG~\cite{LPIPS} for low dynamic range (LDR) images. On the Stanford-ORB benchmark, we also compute PSNR for high dynamic range (HDR) images, denoted as PSNR-H, while PSNR for LDR images is referred to as PSNR-L. For methods that only produce LDR images, we apply the inverse process of the sRGB tone mapping to transform the outputs into linear values. 
To address ambiguities in the relighting task, we align the predicted outputs with the ground truth images by applying channel-wise scale factors prior to metric computation. For Stanford-ORB, these scale factors are determined separately for each output image. For TensoIR, a single global scale factor is calculated and uniformly applied to all output images.

\section{Social Impacts}
Our work presents an algorithm for photorealistic object relighting, which can provide an immersive experience for VR/AR products. It has the potential to simplify labor-intensive workflows in 3D content creation and empower artists in the shopping, film, and gaming industries. It can also be used to augment 3D/multi-view datasets with diverse lighting conditions, potentially benefiting downstream tasks that rely on large-scale photorealistic 3D/multi-view training data. However, this technology also carries potential risks. Misuse of the relighting capability might enable the creation of fraudulent or harmful visual content. Additionally,  our diffusion models require significant computational resources for training, which could bring environmental concerns due to high electricity consumption.


\end{document}